\documentclass[a4paper]{article}
\usepackage{multirow}
\usepackage{INTERSPEECH2022}
\usepackage{caption}
\usepackage{subcaption}
\usepackage{amsmath}
\usepackage{url}
\usepackage{tikz}
\usepackage{pgfplots}
\pgfplotsset{compat=1.17}
\title{Two-Pass Low Latency End-to-End Spoken Language Understanding}
\name{Siddhant Arora, Siddharth Dalmia, Xuankai Chang, Brian Yan, Alan Black, Shinji Watanabe}
\address{Carnegie Mellon University}
\email{\{siddhana, sdalmia, xuankaic, byan, awb, swatanab\}@andrew.cmu.edu}

\begin{document}

\maketitle
\begin{abstract}
  End-to-end (E2E) models are becoming increasingly popular for spoken language understanding (SLU) systems and are beginning to achieve competitive performance to pipeline-based approaches. However, recent work has shown that these models struggle to generalize to new phrasings for the same intent indicating that models cannot understand the semantic content of the given utterance. In this work, we incorporated language models pre-trained on unlabeled text data inside E2E-SLU frameworks to build strong semantic representations. Incorporating both semantic and acoustic information can increase the inference time, leading to high latency when deployed for applications like voice assistants. We developed a 2-pass SLU system that makes low latency prediction using acoustic information from the few seconds of the audio in the first pass and makes higher quality prediction in the second pass by combining semantic and acoustic representations. We take inspiration from prior work on 2-pass end-to-end speech recognition systems that attends on both audio and first-pass hypothesis using a deliberation network. The proposed 2-pass SLU system outperforms the acoustic-based SLU model on the Fluent Speech Commands Challenge Set and SLURP dataset and reduces latency, thus improving user experience. Our code and models are publicly available as part of the ESPnet-SLU toolkit.
\end{abstract}
\section{Introduction}
Spoken Language Understanding (SLU) is an essential component of many daily applications like voice assistants, social bots, and intelligent home devices \cite{socialbot, snips-voice-platform,Personal_assistant}. Conventional SLU systems consist of a pipeline-based model where a speech recognition (ASR) system converts input audio into text, followed by Natural Language Understanding (NLU) system that produces intent from predicted text. This enables utilizing the vast abundance of ASR and NLU research \cite{SLURP} for quick development of SLU systems. However, the cascaded systems suffer from various drawbacks. First, errors in the ASR transcripts can adversely affect the performance of the NLU models. Second, the audio signal consists of non-phonemic signals such as pauses, intonations which can provide additional cues to determine the semantic content of an utterance that a text-based system cannot capture. Thus, many end-to-end SLU (E2E-SLU) models \cite{agrawal2020tie,saxon21_interspeech,ganhotra21_interspeech} have been proposed that can avoid drawbacks of the cascaded systems by identifying intent directly from audio. Further, E2E-SLU models have a smaller carbon footprint \cite{agrawal2020tie} making them particularly attractive for performing on-device SLU. 

However, recent work \cite{FSC_MASE} has shown that E2E-SLU systems struggle to generalize to unique phrasing for the same intent, suggesting an opportunity for enhancing semantic modeling of existing SLU systems. A number of approaches \cite{agrawal2020tie,speechbert,chung2020splat} have been proposed to learn semantic content directly from audio. These approaches aim to incorporate pretrained language models \cite{BERT,MPNet} to improve semantic processing of SLU architectures. Since SLU datasets are expensive to annotate, there is often a small amount of labeled SLU data. Using pretrained language models trained on a large amount of unlabelled text can help to build rich semantic representations. One approach \cite{chung2020splat,huang2020leveraging} to enhance the semantic power of these systems is to jointly train a speech to intent and text to intent model by finetuning a pretrained LM and then align the acoustic and text embeddings in a shared latent space. There has also been an effort \cite{lai2021semi} to concatenate acoustic embeddings and semantic embeddings produced by a large LM from ASR transcript to predict the intent for the given utterance. Some use an interface to join the acoustic encoder with the pretrained NLP model~\cite{seo2021integration}.

Further, a high latency E2E-SLU system can affect the naturalness of human-machine interaction when these systems are deployed in commercial applications like voice assistants. If the SLU models have high latency, it can disrupt human-computer interaction, severely impacting user experience. As a result, some approaches \cite{potdar2021streaming,shivakumar2021rnn} have been recently proposed, which identify the intention of the user with reduced delay.

In this work, we propose an architecture that extends the prior research by building SLU systems that not only model the semantic content of an utterance but also with low latency. We take inspiration from 2-pass ASR systems \cite{hu2020deliberation,sainath2019two} where the first pass model predicts with low latency, and the second pass model improves on the initial prediction from the first pass model. Our first pass model can predict intent only from the first few seconds of audio, whereas the second pass model combines acoustic information from the entire speech and semantic information from ASR-hypothesis using a deliberation network \cite{xia2017deliberation}. Both models share the acoustic encoder to reduce the model size and computation cost. Further, we combine the predictions from the first pass and second pass models by using a threshold over the confidence score of the first pass model. This reduces the latency of our system further as we use the second pass model prediction only when the first pass model is not confident about its prediction. Our experiments are conducted using the Fluent Speech Command (FSC) challenge \cite{FSC_MASE,Lugosch_FSC} test set and Spoken Language Understanding Resource Package (SLURP) \cite{SLURP} dataset. Our proposed two-pass low-latency E2E SLU model achieves a 2.1\% absolute improvement over the only acoustic-based SLU system on intent classification in the FSC Challenge Utterance Set while being just as fast as the single pass model. Our code and models are publicly available as part of the ESPNet-SLU \cite{ESPnet-SLU} toolkit \footnote{The interactive demo showing our 2-pass SLU model identifying intent is available at - https://espnet-slu.github.io/two-pass}.
 
The key contributions of our work are summarised below-
\begin{itemize}
    \item We propose a 2-pass SLU system where the second pass model can use both acoustic embeddings from the first pass encoder and semantic embedding of the ASR hypothesis generated from a pretrained LM. We can show that our 2-pass SLU system can improve intent classification, particularly for unseen word phrasings.
    \item 
    We observe that first pass system can achieve decent intent classification accuracy with reduced latency by looking at only the first few seconds of audio. By running the first pass model only when it's confidence is high, we can reduce the overall inference time.
\end{itemize}

\section{DELIBERATION TWO-PASS E2E SLU}
\begin{figure}[t]
  \centering
    \includegraphics[width=\linewidth]{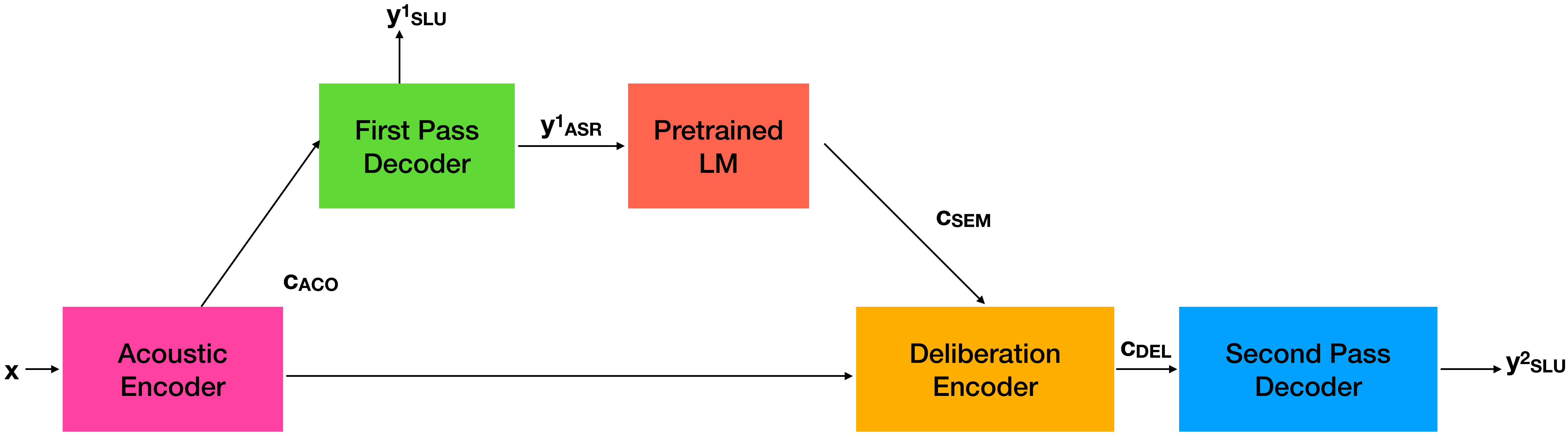}
  \caption{Diagram of Deliberation model based 2-pass SLU system where the $2^{nd}$ pass attends both on acoustic information from $1^{st}$ pass and semantic information from ASR transcript.}
  \label{img:Semantic_SLU}
  \vskip -0.2in
\end{figure}
\subsection{Model architecture}
The architecture of our 2-pass SLU system is shown in Figure ~\ref{img:Semantic_SLU}. The input to our system is $X = (\mathbf{x}_t \in \mathcal{R}^d | t =1,...,T)$, a T-length audio feature sequence with feature dimension
size D. This input speech signal is passed through an acoustic encoder ($\text{ENC}_{\text{aco}}$) to generate acoustic embeddings ($\textbf{c}_{\text{aco}}$),
\begin{equation}
    \textbf{c}_{\text{aco}}= \text{ENC}_{\text{aco}}(X)
\end{equation}
This acoustic embedding is a sequence of latent representations $\textbf{c}_{\text{aco}}=(\textbf{c}_{\text{aco}}^{1},\textbf{c}_{\text{aco}}^{t},..,\textbf{c}_{\text{aco}}^{T})$ where $\textbf{c}_{\text{aco}}^{t} \in \mathcal{R}^d$.
The first pass decoder $\text{DEC}^{1}$ then maps the acoustic embedding $\textbf{c}_{\text{aco}}$ and preceding tokens output by the first pass decoder $\hat{y}^{1}_{1:L-1}$ to $\textbf{h}_{l}^{D}$.
\begin{equation}
    \textbf{h}_{l}^{D}= \text{DEC}^{1}(\textbf{c}_{\text{aco}},\hat{y}^{1}_{1:L-1})
\end{equation}
The likelihood of each token in the output sequence is given by $\text{SOFTMAXOUT}^{1}$ which denotes a linear layer that maps decoder output $\textbf{h}_{l}^{D}$ to vocabulary $\mathcal{V}$ followed by softmax function.
\begin{equation}
    P(\hat{y}_{l}^{1}|\textbf{c}_{\text{aco}},\hat{y}^{1}_{1:L-1})= \text{SOFTMAXOUT}^{1}(\textbf{h}_{l}^{D})
\end{equation}
The likelihood of the entire sequence is computed by composing the conditional probabilities at each step for L tokens,
\begin{equation}
    P(\hat{y}^{1}|\textbf{c}_{\text{aco}})= \Pi_{i=1}^{L} P(\hat{y}_{i}^{1}|\textbf{c}_{\text{aco}},\hat{y}^{1}_{1:i-1})
\label{conf_first_phase}
\end{equation}
In our system, we train with an auxiliary ASR objective~\cite{ESPnet-SLU,deoras2012joint} by making the decoder generate both the intent ($y^{1}_{\text{slu}}$) and ASR transcript ($y^{1}_{\text{asr}}$). Thus the output sequence $y^{1}$ consist of intent followed by ASR transcript ($y^{1}_{\text{slu}}||y^{1}_{\text{asr}}$). This output sequence is computed by maximising the conditional likelihood of entire sequence from output sequence $\mathcal{S}$ of size $\|\mathcal{V}\|^{L}$.
\begin{equation}
    y^{1}_{\text{slu}}||y^{1}_{\text{asr}} = \operatorname*{argmax}_{\hat{y}^{1} \in S} P(\hat{y}^{1}|\textbf{c}_{\text{aco}})
\label{first_phase}
\end{equation}
The ASR transcript is first tokenized into sequence of length $\hat{T}$ then passed to a pretrained LM ($\text{ENC}_{\text{lm}}$) to generate semantic representations 
$\hat{\textbf{c}}_{\text{sem}} =(\hat{\textbf{c}}_{\text{sem}}^{1},\hat{\textbf{c}}_{\text{sem}}^{t},..,\hat{\textbf{c}}_{\text{sem}}^{\hat{T}})$
where $\hat{\textbf{c}}_{\text{sem}}^{t} \in \mathcal{R}^o$ ,
\begin{equation}
    \hat{\textbf{c}}_{\text{sem}} = \text{ENC}_{\text{lm}}(y^{1}_{\text{asr}})
\label{semantic}
\end{equation}
The semantic ($c_{\text{sem}}$) embeddings are then passed to linear layer $\mathcal{M}^{o*d}$ to produce representations with the same hidden dimension as $\textbf{c}_{\text{aco}}$ i.e. $\textbf{c}_{\text{sem}} = (\textbf{c}_{\text{sem}}^{1},\textbf{c}_{\text{sem}}^{t},..,\textbf{c}_{\text{sem}}^{\hat{T}})$ where $\textbf{c}_{\text{sem}}^{t} \in \mathcal{R}^d$
\begin{equation}
    \textbf{c}_{\text{sem}} = \mathcal{M}^{o*d}(\hat{\textbf{c}}_{\text{sem}})
\end{equation}
The acoustic and semantic embeddings are then concatenated together to produce a sequence of length $T+\hat{T}$  and are then attended by a deliberation encoder to produce joint embedding ($\textbf{c}_{\text{del}}$).
\begin{equation}
    \textbf{c}_{\text{del}} = \text{ENC}_{\text{del}}(\textbf{c}_{\text{aco}}|| \textbf{c}_{\text{sem}})
\label{deliberation_enc}
\end{equation}
The second pass decoder $\text{DEC}^{1}$ maps this joint embedding $\textbf{c}_{\text{del}}$ and previous tokens output by the second decoder $\hat{y}^{2}_{1:L-1}$ to $\hat{\textbf{h}}_{l}^{D}$.
\begin{equation}
    \hat{\textbf{h}}_{l}^{D}= \text{DEC}^{2}(\textbf{c}_{\text{del}},\hat{y}^{2}_{1:L-1})
\label{second_pass_decoder}
\end{equation}
 Further similar to the first pass decoder, we can train with an auxiliary ASR objective as shown in the equations below.
\begin{equation}
    P(\hat{y}_{l}^{2}|\textbf{c}_{\text{del}},\hat{y}^{2}_{1:L-1})= \text{SOFTMAXOUT}^{2}(\hat{\textbf{h}}_{l}^{D})
\end{equation}
\begin{equation}
    P(\hat{y}^{2}|\textbf{c}_{\text{del}})= \prod_{i=1}^{L} P(\hat{y}_{i}^{2}|\textbf{c}_{\text{del}},\hat{y}^{2}_{1:i-1})
\end{equation}
\begin{equation}
    y^{2}_{\text{slu}}||y^{2}_{\text{asr}} = \operatorname*{argmax}_{\hat{y}^{2} \in S} P(\hat{y}^{2}|\textbf{c}_{\text{del}})
\label{second_phase}
\end{equation}
There are two main differences between our approach and approaches used in prior work. Previous work \cite{lai2021semi} has tried to combine acoustic and semantic information by simply concatenating acoustic and semantic embeddings after bringing them to the same length in token dimension. In this work, we attend to both semantic ($\textbf{c}_{\text{sem}}$) and acoustic ($\textbf{c}_{\text{aco}}$) embedding to compute a joint embedding ($\textbf{c}_{\text{del}}$) as shown in equation \ref{deliberation_enc}. This procedure helps us to better recover from errors in ASR transcripts by attending more towards acoustic embeddings in such cases (See section \ref{sec:heatmap}). Further, our methods extract intent in two phases, using only acoustic information in the first phase with low latency as described in equation \ref{first_phase} and then combining audio and semantic information in the second phase as described in equation \ref{second_phase}. This 2 phase inference setup reduces our system's overall latency. Further, both these phases use a shared acoustic encoder ($\text{ENC}_{\text{aco}}$ in equation 1), thus reducing the carbon footprint of 2-pass SLU architecture.
\subsection{Training}
\label{sec:training}
Similar to prior work ~\cite{xia2017deliberation} on 2-pass ASR systems, we used a two step training process for our SLU system. In the first step, we train only the acoustic encoder ($\text{ENC}_{\text{aco}}$ in equation 1) and first-pass decoder ($\text{DEC}^{1}$ in equation 2) to train the acoustic modeling of our SLU architecture. In the second step, the model learns to combine semantic and acoustic information by training deliberation encoder ($\text{ENC}_{\text{del}}$ in equation \ref{deliberation_enc}) and second pass decoder ($\text{DEC}^{2}$ in equation \ref{second_pass_decoder}). We do not use teacher forcing during the second phase, and the model is trained using ASR transcripts. We can also finetune the pretrained LM, but we did not observe significant gains in our initial investigations.
\subsection{Decoding}
Our decoding also consists of two phases - firstly, we get the intent ($y^{1}_{\text{slu}}$) and ASR transcript ($y^{1}_{\text{asr}}$) from the first pass decoder, as in equation \ref{first_phase} and secondly, we attend to both ASR transcript ($y^{1}_{\text{asr}}$) and acoustic embedding ($\textbf{c}_{\text{aco}}$) to get second pass intent ($y^{2}_{\text{slu}}$), as in equation \ref{second_phase}. We use beam search to find the output sequence that maximizes conditional likelihood during both phases of inference. Note that in some cases, the dataset might be too small to learn reliable ASR transcripts, and hence we also experimented with pretrained ASR models for generating these transcripts. Finally, we also investigated reducing inference time of our first pass SLU model by predicting intent from only the first few seconds of audio.

\section{Experiment Setup}

\subsection{Dataset}
We evaluated our proposed approach on publicly available SLU datasets, namely FSC \cite{Lugosch_FSC} and SLURP \cite{SLURP} dataset. FSC is a free, reasonably large SLU dataset that contains around 30K utterances and includes 31 intent types. 
For the FSC dataset, we used the Challenge \cite{FSC_MASE} test split, which consists of 2 test sets: Challenge Speaker Set with held out speakers and Challenge Utterance Set with held out utterance transcripts. Challenge Utterance Set serves as an interesting testbed to evaluate the ability of our proposed methodology to improve the semantic generalization to unseen word phrasings of existing SLU architectures.
SLURP contains 72K real and 69K synthetic utterances of single-turn user conversation with a home assistant making it closer to real-life application scenarios. It includes 69 intent types and 56 unique entities and has much higher lexical and semantic diversity than most publicly available SLU datasets. 

\subsection{Architecture details and training}
Our models are implemented in pytorch \cite{pytorch}
and the experiments are conducted through the ESPNet-SLU \cite{ESPnet-SLU} toolkit. We compared our approach using strong speech only baselines consisting of conformer~\cite{conformer,conformer_another} encoder and transformer~\cite{transformer,transformer_another} decoder that are available in  ESPNet-SLU toolkit and achieve near SOTA performance. We also compared our performance to the SOTA NLU results reported on these benchmarks \cite{FSC_MASE,SpeechBrain}. To train a 2-pass system, we used the same acoustic encoder and first pass decoder as our baseline systems. We generated ASR transcript using the first pass decoder for SLURP dataset. Since FSC dataset is too small to learn reliable ASR system, we used a pretrained ASR model trained on Gigaspeech \cite{gigaspeech} dataset to compute ASR transcripts. To get pretrained LMs as semantic encoder, we incorporated the usage of HuggingFace Transformers library \cite{wolf-etal-2020-transformers}, which provides numerous generic and task-specific pretrained language models. The semantic embeddings ($\textbf{c}_{\text{sem}}$ in equation \ref{semantic}) are generated through the output of the last encoder layer of pretrained LM. In this work, we have only experimented with BERT \cite{BERT} as pretrained LM; however, our codebase can incorporate any of the NLU models provided in the Hugging Face repository.\footnote{https://huggingface.co/models} We experimented with both conformer and transformer based architectures for our deliberation encoder while decoder is 6 layer transformer network. During training, SpecAugment \cite{specaugment} was performed for data augmentation. We applied dropout \cite{dropout} and label smoothing \cite{label-smoothing} to mitigate overfitting. 
More details about our models will be publicly available in ESPnet.

\section{Results}

\subsection{Result on entire audio}
\begin{table}[t]
\caption{Intent Classification accuracy
on FSC Challenge Set\cite{Lugosch_FSC,FSC_MASE} for models using only speech, both speech and ASR transcript and speech and ground text. 
}
 \centering
  \resizebox {0.9\linewidth} {!} {
\begin{tabular}{l|l|cc}
\toprule
& Model   & \multicolumn{2}{c}{IC (\% Acc)}                                                  \\
& & Utt Test & Spk Test \\\midrule
\multirow{1}{*}{\shortstack[l]{Baseline}}& E2E-SLU \cite{ESPnet-SLU}                                         & 78.5 & 97.5 \\ \midrule
\multirow{2}{*}{\textbf{SLU w/ Semantics}} & Transformer Deliberation & 82.3 & 98.1 \\
& Conformer Deliberation & 81.9 & 97.5\\ \midrule
\multirow{3}{*}{\shortstack[l]{Oracle Text}}
& Only Text \cite{FSC_MASE}   & 86.1 & 100 \\
& Transformer Deliberation  & 87.6 & 100 \\
\bottomrule
\end{tabular}
}
\label{tbl:FSC}
\vskip -0.03in
\end{table}
\begin{table}[t]
\caption{Intent Classification accuracy
on SLURP \cite{SLURP} for models using only speech, both speech and ASR transcript and speech and ground text. 
}
 \centering
  \resizebox {0.9\linewidth} {!} {
\begin{tabular}{l|l|c}
\toprule
& Model                                                             & IC (\% Acc)  \\\midrule
\multirow{1}{*}{\shortstack[l]{Baseline}}& E2E-SLU \cite{ESPnet-SLU}                                         & 86.3 \\ \midrule
\multirow{2}{*}{\textbf{SLU w/ Semantics}} & Transformer Deliberation & 86.4 \\
& Conformer Deliberation & 86.6 \\
\midrule
\multirow{3}{*}{\shortstack[l]{Oracle Text}}
& Only Text \cite{SpeechBrain}   & 87.3 \\
& Conformer Deliberation & 89.0 \\
\bottomrule
\end{tabular}
}
\label{tbl:SLURP}
\vskip -0.15in
\end{table}
Table \ref{tbl:FSC} shows that our proposed 2-pass SLU system outperforms the only acoustic based SLU system both on Challenge Utterance and Speaker Test Set. We observe an absolute 3.8\% improvement for the utterance set, indicating that our approach is better at generalizing to unique phrasings of an intent. This provides evidence that pretrained LM can be used to enhance the semantic processing of SLU systems and can help bridge the performance gap between SLU and NLU architectures.

We further investigate running our 2-pass system using ground truth text and report that our approach outperforms the text-only performance. Based on these results, we hypothesize that speech contains non phonemic signals like pauses, intonations that can help improve performance over text-only systems. This finding illustrates that our approach of jointly encoding speech and text in a unified framework is more promising than a pipeline-based architecture.

We further conducted our experiments on SLURP dataset and report similar findings in Table \ref{tbl:SLURP}. We observed gains on combining acoustic and semantic information however the gains are smaller since the gap between SLU and NLU system is smaller for this benchmark. This shows that our method can be extended to improve performance on other SLU datasets.
\subsection{Result on first x seconds of audio}

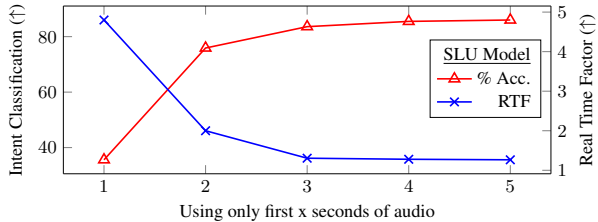
\begin{figure}[t]
\resizebox {\linewidth} {!} {

\begin{tikzpicture}
	\begin{axis}[
		xlabel=Using only first x seconds of audio,
		ylabel=Intent Classification ($\uparrow$),
		axis y line*=left,
		xtick=data,
        width=10cm,
        height=4.5cm,
	    every axis plot/.append style={thick},
		]
	\addplot[color=red,mark=triangle, mark options={scale=1.5}] coordinates {
	    (1, 35.6)
	    (2, 75.9)
	    (3, 83.6)
	    (4, 85.5)
	    (5, 86.0)
	}; \label{plot_one}
	\end{axis}
	
	\begin{axis}[
		axis y line*=right,
		axis x line=none,
		xtick=data,
		ylabel=Real Time Factor ($\uparrow$),
		legend cell align={right},
		legend style={at={(0.98,0.80)},anchor=north east},
        width=10cm,
        height=4.5cm,
	    every axis plot/.append style={thick},
		]
		\addlegendimage{empty legend}\addlegendentry{\hspace{-.6cm}\underline{SLU Model}}
		\addlegendimage{/pgfplots/refstyle=plot_one}\addlegendentry{\% Acc.}
	\addplot[color=blue,mark=x, mark options={scale=1.5}] coordinates {
		(1, 4.8)
	    (2, 2)
	    (3, 1.3053)
	    (4, 1.27835)
	    (5, 1.2653)
	}; \addlegendentry{RTF}
	\end{axis}
\end{tikzpicture}
}
  \caption{Intent classification accuracy using the first 'x' seconds of audio against the real time factor (compared to full audio) on the SLURP dataset. Results for our acoustic based SLU model using only the first 1\dots5 seconds of audio.}
  \label{fig:beam_exp}
\vskip -0.25in
\end{figure}
\begin{table}[!b]
\caption{Intent Classification 
on FSC Challenge Set for first and second pass model based on the confidence of first pass model
}
 \centering
  \resizebox {\linewidth} {!} {
\begin{tabular}{l|cc|l|cc}
\toprule
Confidence Score & \multicolumn{2}{|c|}{Support} & Model & \multicolumn{2}{c}{IC (\% Acc)}                                                  \\
& Utt Test & Spk Test & & Utt Test & Spk Test \\\midrule
\multirow{2}{*}{$>80\%$} & \multirow{2}{*}{1604} & \multirow{2}{*}{2694} & First Pass & 84.9 & 97.6 \\
 & & & Second Pass & 89.5 & 99.2 \\
\midrule
\multirow{2}{*}{$<80\%$} & \multirow{2}{*}{2600} & \multirow{2}{*}{655} & First Pass & 61.8 & 63.0 \\
& & & Second Pass & 77.9 & 93.6 \\
\bottomrule
\end{tabular}
}
\label{tbl:Conf_FSC}
\vskip -0.15in
\end{table}
We investigated the system performance on predicting intent only from the first few seconds of audio. We hypothesize that sufficient evidence can be gathered from the first few seconds of audio to extract intent for the majority of utterances, and we did not need to wait till the end of the utterance. Our experiments shown in Figure \ref{fig:beam_exp} demonstrate that we can achieve decent accuracy while making significant reductions in the inference time per utterance by using only the first 2 seconds of audio. Our final architecture uses the only speech model that predicts using the first 2 seconds as our first pass SLU system and the speech+transcript model that predicts using the entire audio as our second pass SLU system.

\subsection{Combining first pass and second pass predictions}

For the FSC Challenge test set, the inference time of the first pass model is nearly 75\% of that of the second pass model. 
 However by observing the confidence score for first pass model predictions as computed in equation \ref{conf_first_phase}. We found that a majority of performance gains from the second pass model occur when the first pass model is not confident about it's predictions, as shown in Table \ref{tbl:Conf_FSC}. Motivated by this observation, we use the second pass model only when the first pass model is not confident (under a certain threshold) about its predictions to reduce the overall latency of our SLU system. Through this setup, we were able to improve on the only acoustic-based SLU system by achieving 80.6\% accuracy while also predicting on average 90\% faster than the second pass SLU system for FSC Utterance Test Set. Our observations were similar for FSC Speaker Test Set and SLURP dataset as shown in Table \ref{tbl:Conf_FSC} and Table \ref{tbl:Conf_SLURP}.
 \begin{table}[!b]
\caption{Intent Classification 
on SLURP dataset for first and second pass model based on the confidence of first pass model
}
 \centering
   \resizebox {0.9\linewidth} {!} {
\begin{tabular}{l|c|l|c}
\toprule
Confidence Score & Support & Model & Test IC (\% Acc)\\
\midrule
\multirow{2}{*}{$>65\%$} & \multirow{2}{*}{2556} & First Pass & 96.6 \\
 & & Second Pass & 97.4 \\
\midrule
\multirow{2}{*}{$<65\%$} & \multirow{2}{*}{10522} & First Pass & 70.9 \\
& & Second Pass & 84.0 \\
\bottomrule
\end{tabular}
}
\label{tbl:Conf_SLURP}
\vskip -0.15in
\end{table}

\section{Analysis}
\subsection{Breakdown of result based on transcript WER}
\begin{figure}
\centering
\begin{subfigure}{.5\linewidth}
  \centering
  \includegraphics[width=.95\linewidth]{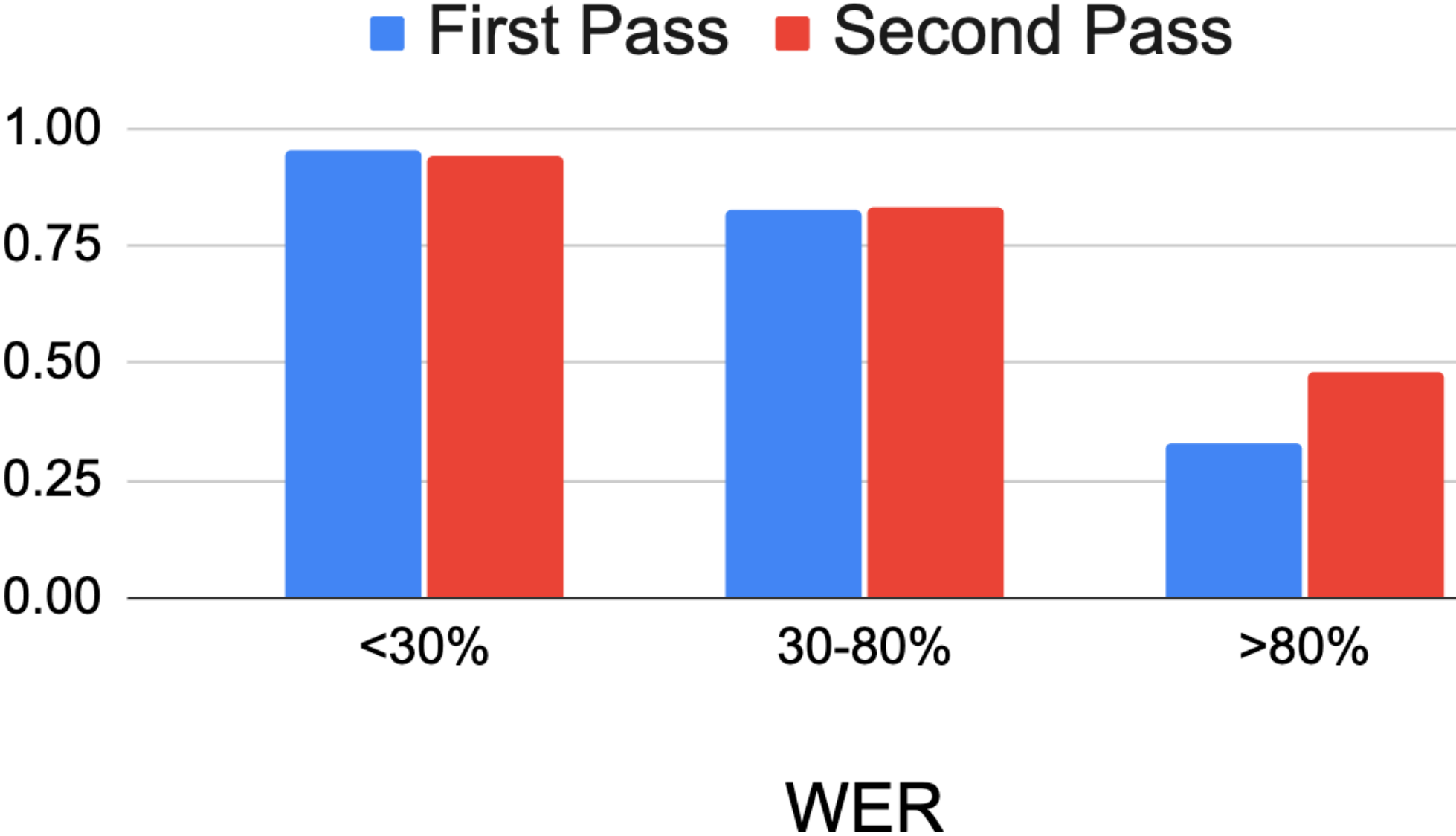}
  \caption{SLURP}
  \label{fig:WER_Slurp}
\end{subfigure}%
\begin{subfigure}{.5\linewidth}
  \centering
  \includegraphics[width=.95\linewidth]{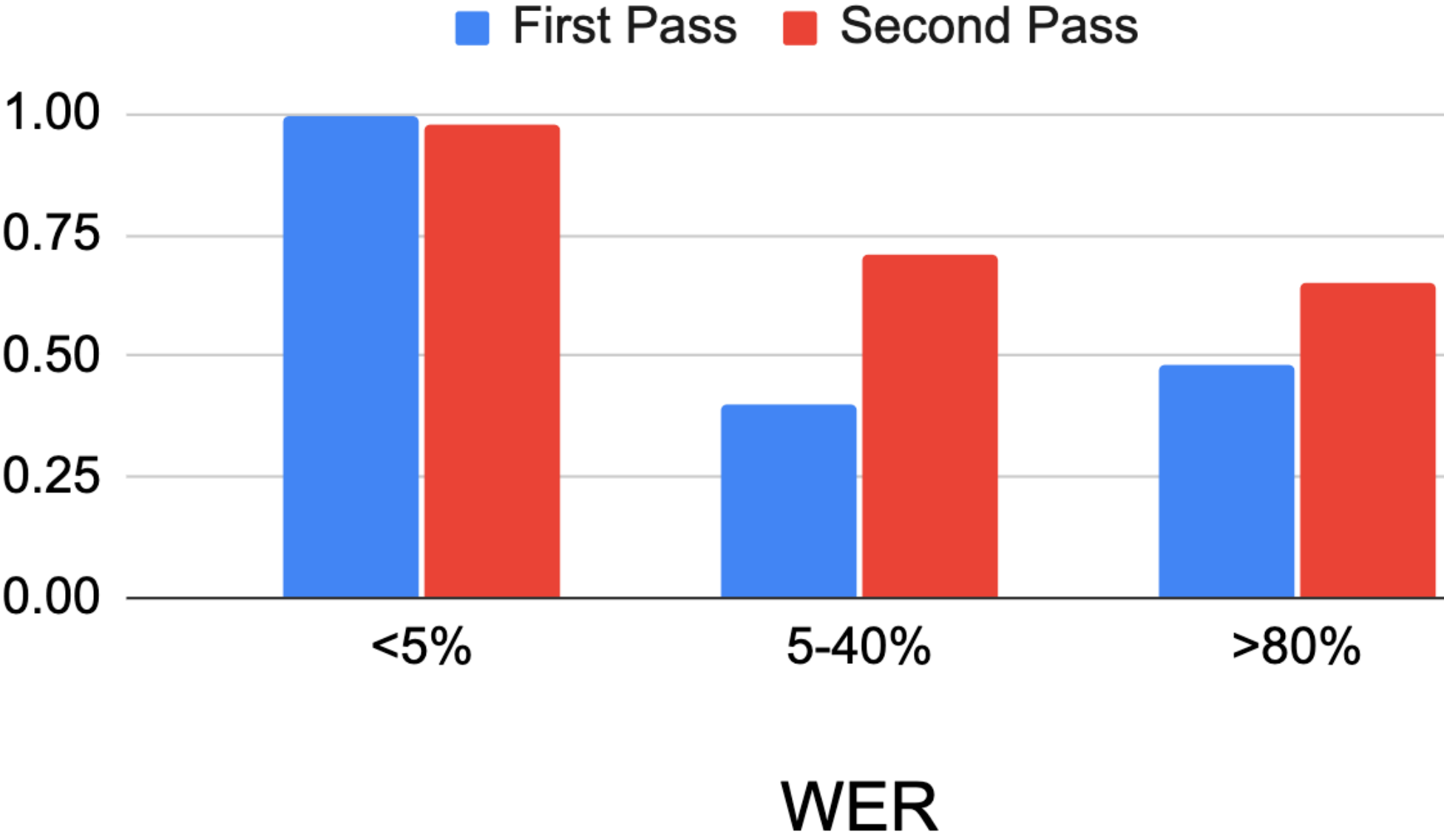}
  \caption{FSC Unseen Utterance Set}
  \label{fig:WER_FSC_Utt}
\end{subfigure}
\caption{First Pass and Second Pass Intent classification accuracy against the WER of the predicted transcript.}
\label{fig:WER}
\vskip -0.15in
\end{figure}

We compared the first pass and second pass SLU accuracy based on WER of ASR transcript generated by first-pass model through auxiliary ASR objective. Our findings for FSC Challenge Utterance Set are shown in Figure \ref{fig:WER_FSC_Utt}. We observed that first pass and second pass SLU models have very similar accuracy when the WER of first-pass ASR transcript is lower than 5\%, making nearly 47\% of spoken utterances in Unseen Utterance Set. However, the second pass model improved performance significantly when the WER of the first pass ASR transcript was high. We observe similar findings for the SLURP dataset as shown in Figure \ref{fig:WER_Slurp}.
After a fine grained analysis, we inferred that while some of these cases belong to spoken utterances that had intent relevant information in the later part of audio, other cases of performance difference can be attributed to better semantic modeling, which even helped the model to recover from errors in ASR transcript. For instance, for a given spoken utterance in the SLURP dataset that says ``please tell me the score of the game'', the first pass model incorrectly predicts the intent as ``PLAY\_GAME'' and transcript as ``please tell me the sports game''. The second pass SLU system can then attend both on ASR transcript and audio file to correctly predict the intent as ``GENERAL\_QUIRKY''. We hypothesize that this is because the words ``tell me the score of the game'' make better semantic sense, and hence the model was able to infer the intent as a general quirky query.

\subsection{Attention heatmaps}
\label{sec:heatmap}
\begin{figure}[tbp]
  \centering
    \includegraphics[width=0.6\linewidth]{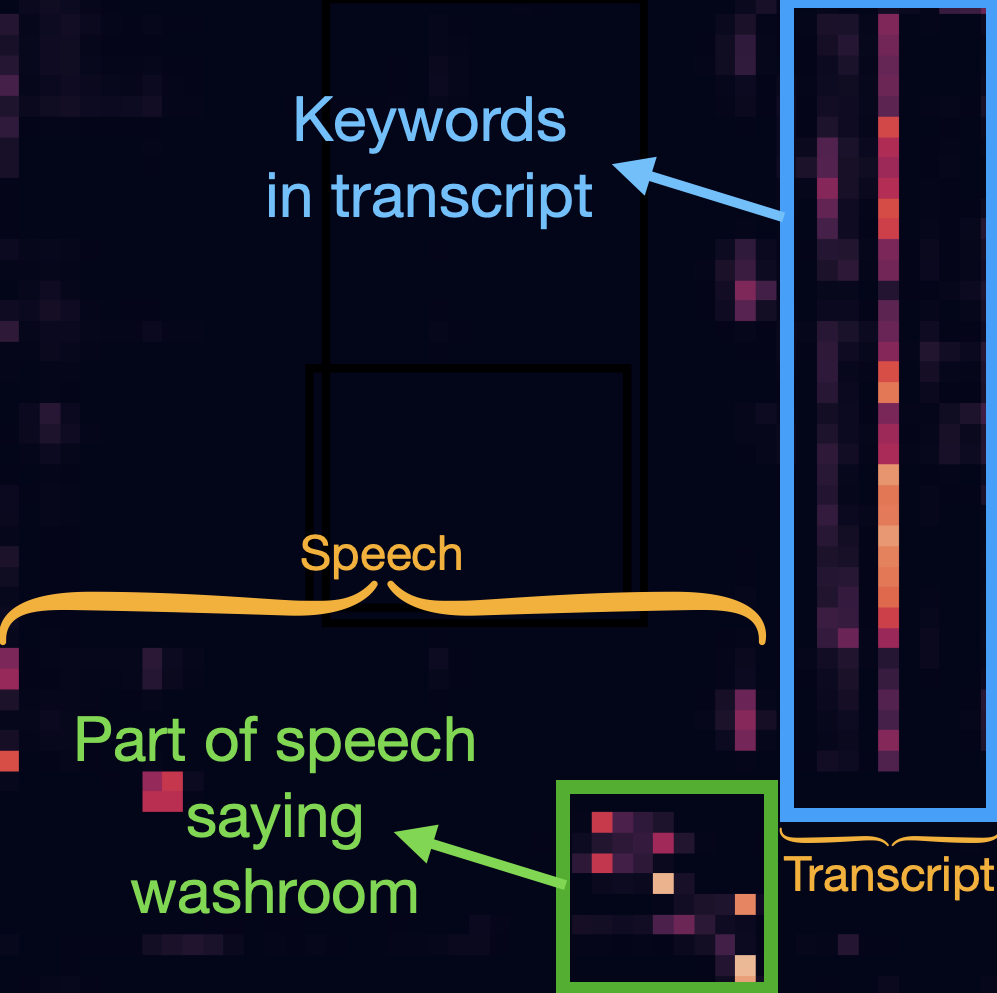}
  \caption{Sample heatmap of the first deliberation encoder layer}
  \label{img:Attention_Heatmap}
  \vskip -0.25in
\end{figure}

We also plot the attention heatmaps to illustrate that our proposed system can recover from ASR errors. Figure \ref{img:Attention_Heatmap} shows a sample attention map for a spoken utterance in FSC dataset, which says ``increase the heating in the washroom''. The ASR transcript that the second pass model attends to reads, ``increase the heating in the usroom''. We observe the model attends more to keywords in transcript like ``increase'' and ``heating''. Further, it attends to part of speech that says ``washroom'' to recover from the error in the ASR transcript.

\section{Conclusions}
We propose a novel 2-pass SLU system that combines semantic and acoustic information to achieve higher quality prediction while reducing the latency of the SLU system. We show evidence that our approach can improve performance on the existing spoken language understanding benchmarks, particularly in generalization to new phrasings for the same intent. We further gain an understanding of the reasons for the improvement in the performance of our second pass SLU system and observe that our approach can correctly extract intent from utterances that are challenging for ASR systems. As our methodology is task-agnostic, we aim to explore the extension of our proposed architecture to other SLU tasks, such as entity classification as well as other speech processing tasks like speech translation. 
\section{Acknowledgement}
This work used the Extreme Science and Engineering Discovery Environment (XSEDE) \cite{xsede}, which is supported by NSF grant number ACI-1548562. Specifically, it used the Bridges system \cite{nystrom2015bridges}, which is supported by NSF award number ACI-1445606, at the Pittsburgh Supercomputing Center (PSC).

\bibliographystyle{IEEEtran}

\bibliography{mybib}

\end{document}